\documentclass[letterpaper, 10 pt, conference]{ieeeconf}  

\IEEEoverridecommandlockouts                              

\usepackage[pagebackref,breaklinks,colorlinks]{hyperref}
\usepackage{multirow}
\usepackage{graphicx} 
\usepackage{float}
\usepackage{pgfplots}
\usepackage{csvsimple}
\usepackage{caption}
\usepackage{subcaption}
\usepackage{hyperref}
\usepackage{booktabs}
\usepackage{amssymb}
\usepackage{comment}
\usepackage{cuted}
\newcommand{\cmark}{\ensuremath{\checkmark}}

\usepackage{pifont}

\title{\LARGE \bf
A2RL V\textsubscript{max}: The A2RL autonomous racing dataset for long-range, high-speed perception and multi-vehicle interaction}

\author{
    Marvin Klemp$^{1,10}$, 
    Dominic Ebner$^{2,11}$,
    Cornelius Schröder$^{2,11}$,
    Davide Malvezzi$^{4}$, 
    László Turányi$^{5}$, \\ 
    Riccardo Donati$^{6}$,  
    Ilia Schminik$^{7}$,
    Xia Ning$^{8}$,
    Yanxin Zhou$^{9}$,
    Matthew Flagg$^{10}$, 
    Christoph Stiller$^{1}$, \\
    Markus Lienkamp$^{2,11}$, 
    Marko Bertogna$^{4}$,
    Gergely Bári$^{5}$,
    Andreas Birk$^{7}$,
    Ren Jin$^{8}$,
    Chen Lv$^{9}$,
    Johannes Betz$^{3,11}$
    \thanks{\small
$^{1}$ Institute of Measurement and Control Systems, Karlsruhe Institute of Technology
$^{2}$ Institute of Automotive Technology, Technical University of Munich
$^{3}$ Professorship of Autonomous Vehicle Systems, Technical University of Munich
$^{4}$ University of Modena and Reggio Emilia
$^{5}$ Humda Lab, Széchenyi István University
$^{6}$ Politecnico di Milano
$^{7}$ Constructor University
$^{8}$ Beijing Institute of Technology
$^{9}$ Nanyang Technological University
$^{10}$ Code 19 Racing
$^{11}$ Munich Institute of Robotics and Machine Intelligence (MIRMI), Technical University of Munich.
\textit{Acknowledgment:} We thank Mathew Arbuckle of Code 19 Racing for supporting the evaluation of existing models on our dataset.}
}

\begin{document}

\maketitle
\thispagestyle{empty}
\pagestyle{empty}


\begin{abstract}

In autonomous driving development, a perception dataset is crucial, as it provides fundamental data for training, testing, and validating algorithms for an autonomous vehicle's multimodal perception systems. So far, most research has concentrated on providing datasets for well-structured urban environments.
This work introduces the A2RL V\textsubscript{max} open-source dataset, specifically designed for perception tasks in high-speed autonomous driving and multi-vehicle interaction. 
The dataset was captured during the 2024 Abu Dhabi Autonomous Racing League (A2RL), held at the Yas Marina F1 Circuit, with participation from all competing teams. 
It contains diverse scenarios, including single-vehicle data at varying speeds, multi-vehicle sessions, and the full final four-vehicle race.
The dataset contains almost 30,000 professionally annotated LiDAR point clouds, along with RADAR point clouds.
In particular, it is the first large-scale dataset in autonomous racing to feature professionally annotated LiDAR point clouds, enabling deep learning-based perception research.
The data is provided in a developer-friendly format, enabling easy implementation and evaluation in future research.
We provide implementation and evaluation for off-the-shelf 3D detection and tracking methods.
Although baseline methods show promising results for both 3D detection and tracking, specialized methods are required to address the unique challenges of high-speed autonomous driving. For a detailed description of the dataset, please visit the \href{https://tum-avs.github.io/A2RL_Dataset_website/}{A2RL V\textsubscript{max} Dataset Website}.

\end{abstract}

\section{INTRODUCTION}
\label{sec:intro}
\bstctlcite{BSTcontrol}

\begin{figure}[h] 
    \centering
    \includegraphics[width=0.95\linewidth]{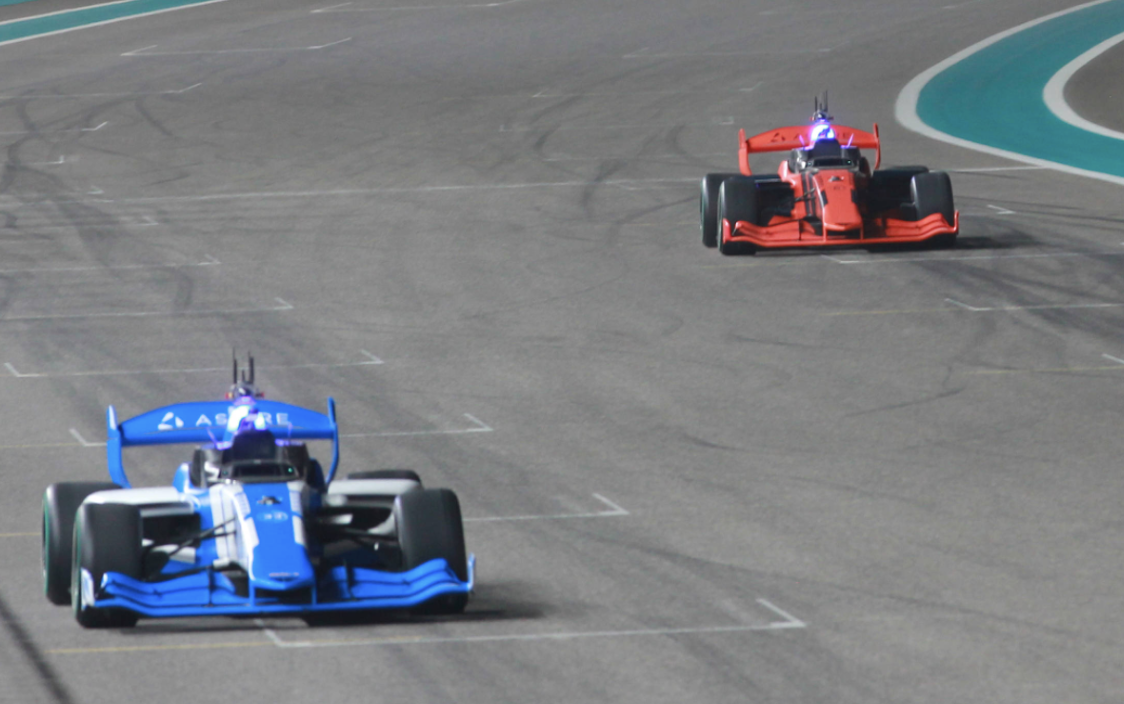}
    \caption{A2RL competition 2024: Head-to-Head race between two autonomous racecars.}
    \label{fig:placeholder}
\end{figure}

The vast amount of research on autonomous vehicles has led to promising results in day-to-day situations. However, managing and providing for rare or atypical situations remains a significant research challenge \cite{visapp19}. Note that those situations are of high interest as autonomous driving accidents tend to occur in the long tail of borderline cases \cite{autonomous-driving-accidents-vs-human-NatureCom24}. 
Although high-speed and high-risk scenarios are rarely encountered on public roads, autonomous systems must be capable of accurately addressing these situations when they do occur. With this in mind, the field of autonomous racing competitions has emerged in recent years \cite{Betz2022}. 

Autonomous racing is an excellent way to investigate such scenarios, since here vehicles can reach high speeds of $> 200$ km/h, which lead to critical conditions for all software, including perception, localization, planning, and control \cite{betz2023fast,Wurman.2022}.
Among these competitions, the F1-Tenth \cite{pmlr-v123-o-kelly20a} challenge has low entry barriers and attracts small groups of university students from all over the world. 
In the Formula SAE Driverless Competition, teams of university students build their own race car from scratch and drive it through a previously unseen track marked by traffic cones \cite{Kabzan.2020}. 

\begin{figure*}[h]
\centering
\includegraphics[width=0.84\textwidth]{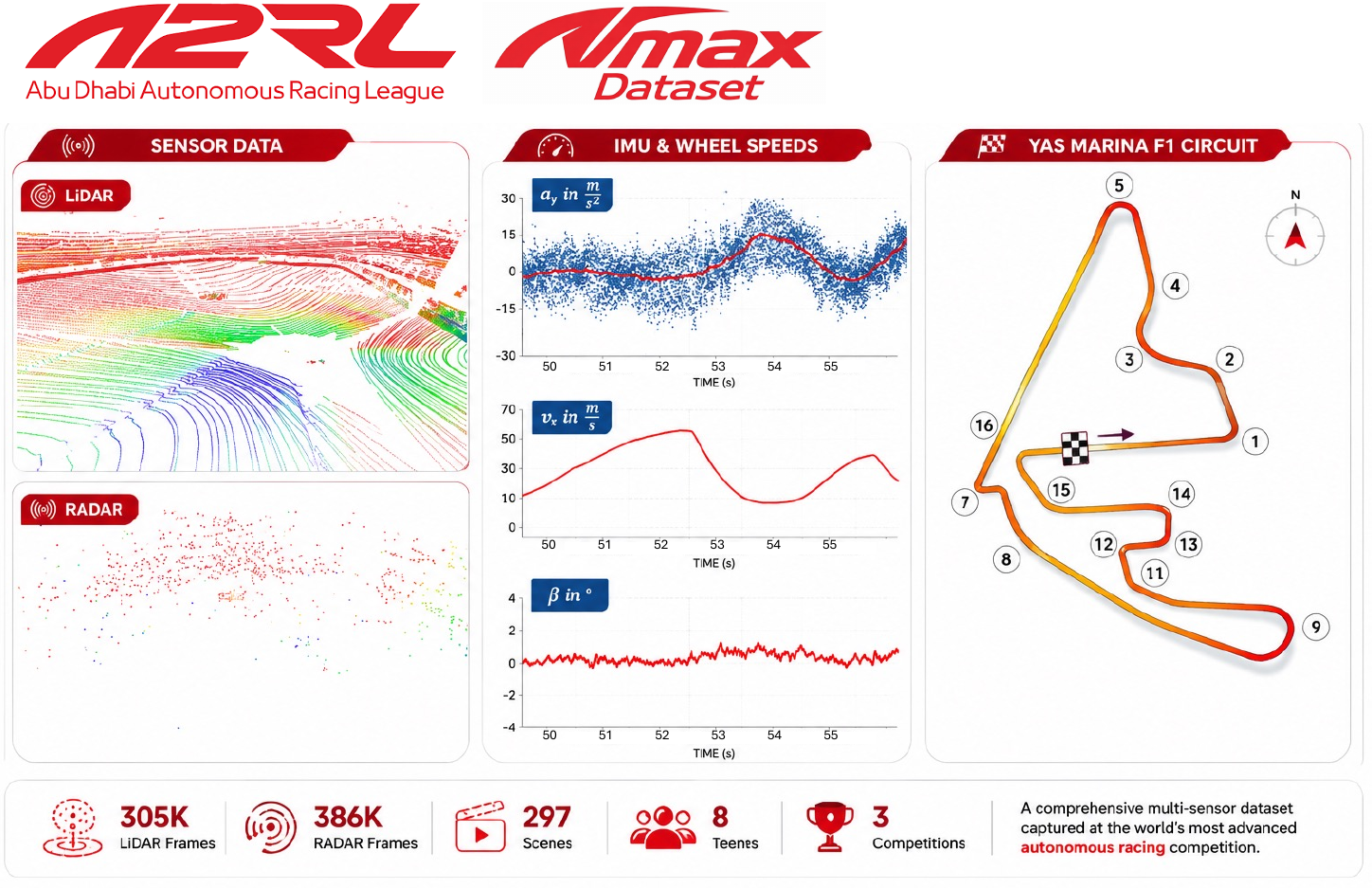}
\caption{A2RL V\textsubscript{max} Dataset Overview: The data was collected over 3 competitions, 8 different teams on one racetrack, and includes sensor Data, vehicle data, and track data.}
\label{fig:main_Image}
\end{figure*}

This paper is enabled by the Abu Dhabi Autonomous Racing League (A2RL), an international high-speed autonomous racing competition \cite{Betz2022_2, Raji.2023, raji2024er, Chung2024, Hoffmann.2026}, organized by ASPIRE, the technology program management arm of the Advanced Technology Research Council (ATRC), Abu Dhabi. By providing access to full-scale race vehicles operating at extreme speeds in competitive multi-agent scenarios, A2RL establishes a unique real-world testbed for long-range perception, multi-vehicle interaction, and system-level validation. The A2RL represents more than a competition format; it is a structured research platform initiated by ASPIRE to accelerate innovation in autonomous mobility under extreme operational constraints. By convening international teams, deploying full-scale race vehicles on a Formula 1 circuit, and fostering cross-team data sharing, A2RL creates a rare, large-scale experimental environment that bridges competitive benchmarking and open scientific collaboration.

To advance research in high-speed, high-interaction autonomous driving, we present the \textit{A2RL V\textsubscript{max} dataset}, captured during the inaugural A2RL competition at the Yas Marina F1 Circuit in 2024 (Figure \ref{fig:main_Image}). The A2RL V\textsubscript{max} dataset reflects this strategic research infrastructure and aims to catalyze high-impact research in safety-critical, high-speed autonomous systems. This dataset enables the evaluation of existing methods under extreme conditions, as well as the development of new algorithms that operate at high speeds with low latency, provide rich perception data in extreme racing conditions, and expand access to autonomous racing research. All eight teams that participated in the A2RL collaborated on this dataset by sharing data, culminating in 28.791 professionally labeled LiDAR point clouds.
The main contributions of this work are the following:
\begin{itemize}
    \item The A2RL V\textsubscript{max} dataset, containing an extensive amount of professionally annotated perception data enabling the use of deep-learning methods as well as their evaluation 
    All materials can be found \href{https://tum-avs.github.io/A2RL_Dataset_website/}{A2RL V\textsubscript{max} Dataset Website}
    \item A reproducible benchmark for 3D object detection and tracking as a reference point for assessing the dataset.
    \item Implementation and evaluation of baseline methods for detection and tracking.
    \item Open source access to the dataset, the source code for the baselines, and the auxiliary material.
    \item Further enrichment of the dataset due to more competitions with more teams within the A2RL competition
\end{itemize}

\section{Related Work}
\label{sec:related_work}

\subsection{Datasets}

In recent years, multiple datasets have been published to advance research in autonomous driving. The KITTI dataset \cite{geiger2013vision} played a fundamental role in facilitating significant progress in object detection and simultaneous localization and mapping. Building on this, larger-scale datasets such as the Waymo Open Perception \cite{Sun.2020}, the nuScenes \cite{nuscenes}, and the Argoverse 2 \cite{Argoverse2} dataset introduced diverse scenarios and richer annotations, enabling the evaluation of multiple perception tasks, including 3D object detection, tracking, prediction, and segmentation. These datasets have emerged as benchmarks in the research of autonomous driving. 
However, they primarily focus on public roads, simple vehicle behavior, and everyday traffic scenarios involving a large number of participants. 
As a result, the speeds of both the recording vehicles and other traffic participants are generally limited to inner-city and highway speeds.

Currently, only the RACECAR dataset \cite{Kulkarni_2023}, collected during the 2021 IAC competitions, is dedicated to high-speed scenarios.
It includes 360° LiDAR, RADAR, camera, and RTK-corrected GNSS data collected on two different oval racetracks in the USA. 
The data were collected at speeds up to 270 km/h. 
However, the fraction of data that includes the interaction between vehicles is small, the number of interacting vehicles is limited to two, and the interactions themselves are subject to a tight set of predefined rules (e.g., coordinated overtaking maneuvers).
Furthermore, the RACECAR dataset does not include professionally annotated data. 
Instead, it provides GNSS positions for both vehicles, which can be used to generate coarse and imprecise labels.
This lack of professionally annotated data makes it challenging for researchers to develop and evaluate algorithms, as coarse, imprecise GNSS-based labels introduce significant uncertainty, reducing the reliability of the ground truth for supervised learning and benchmarking.
 The Formula Student Objects in Context (FSOCO) dataset \cite{fsoco_2022} is a collaborative, contribution-based dataset designed to enhance vision-based cone detection in Formula Student Driverless (FSD) competitions \cite{Strobel2020}. Although the context is autonomous racing, it focuses primarily on relatively slow-speed Formula Student vehicles, with an emphasis on cone detection \cite{Dhall2019}.
Huch et al. \cite{huch2023quantifying} provide a labeled LiDAR point cloud dataset comprising simulated and real-world data for evaluating Sim-to-Real adaptation.
This dataset includes both a fully manually labeled perception dataset recorded during the IAC and a distribution-aligned dataset generated using a LiDAR simulation environment. However, this dataset lacks high-speed driving data, interactive multi-vehicle scenarios, and highly relevant additional sensor data such as radar, IMU, or GNSS. 

\subsection{Perception}

Vehicle detection focuses on accurately detecting and classifying multiple objects in a single scene. 
As a result of the limited range of annotations within datasets, current state-of-the-art 3D detection methods generally aim for high detection accuracy within a maximum range of 50 to 80 meters \cite{geiger2013vision,nuscenes, Sun.2020, OpenPCDet}. 
Among existing datasets, few datasets evaluate detection in longer ranges \cite{fent2024man,Argoverse2,alibeigi2023zenseact,xiao2021pandaset}.
However, the extended detection range poses a significant challenge, as state-of-the-art detectors achieve considerably lower performance on Argoverse 2 than on datasets with an 80-meter limit, with LION achieving a state-of-the-art mAP score of just 0.4 \cite{liu2024lion}.
This suggests that longer detection ranges remain an open challenge in autonomous driving perception, especially since modern LiDAR hardware can return points up to 500 meters \cite{SeyondFalconK}.
Due to the lack of a large-scale dataset containing long-range perception data, research on autonomous racing perception often relies on methods that do not require training data. 
In accordance with the approaches, static points within the point cloud are removed, leaving only moving points (the vehicle) in the point cloud.
These points are then clustered, and a pose estimation algorithm is applied to determine the vehicle's yaw \cite{cellina2025lidarbasedvehicledetectiontracking}.
Furthermore, tracking is significantly more complex since high speeds up to 250 km/h can lead to high differential speeds, such as a position change of several meters within just 50 ms of latency between detections \cite{sauerbeck2023learn, Falanga2019}.
Depending on the LiDAR configuration, latency can be as high as 100 ms, further increasing tracking complexity \cite{Strobel2020}.
A method utilizing an Extended Kalman Filter with a Constant Turn Rate Velocity (CTRV) model to fuse heterogeneous detection inputs from LiDAR, RADAR, and other sensors for robust object tracking in high-speed scenarios is presented in \cite{karle2023multi}.
Although the algorithm is open-source, it has only been evaluated on private data and not on a public dataset.

While relative speeds are high, it is difficult to implement deskewing for detection, since the sensor input cannot be processed for each object individually, even if the speeds of other vehicles are known. Motion-Distortion-Effects have a large impact on static objects, therefore mostly affecting map-based localization, where the map data was recorded at lower velocities.

\section{The A2RL V\textsubscript{max} Dataset}
\label{sec:dataset}

\subsection{The A2RL competition}

The A2RL competition was conceptualized and delivered by ASPIRE as a flagship initiative to advance autonomous systems research at the intersection of artificial intelligence, robotics, and high-performance mobility. Unlike traditional benchmarking efforts, A2RL combines regulatory coordination, infrastructure provision, vehicle standardization, and scientific openness within a single framework. This integrated structure enables systematic data acquisition across multiple teams while preserving competitive integrity.

The A2RL competition was preceded by a two-month development and practice period, bringing together more than 100 participants who had opportunities to use the track for single- or multi-vehicle practice. The A2RL competition consisted of three individual parts: (1) A time-trial, where single vehicles went out to achieve the fastest autonomous lap times. (2) An attack-and-defend session, in which two vehicles engaged in an autonomous head-to-head race, with the goal of scoring as many overtakes as possible. (3) An 8-lap race with four vehicles, where the goal was to race fully autonomously, similar in format to Formula 1 races but without human drivers.
The perception and vehicle data presented in this paper were gathered along these individual competitions.

\begin{table*}[ht]
\centering
\caption{Depending on the team, different data is provided. A tick or a frequency indicates that the specific sensor is available for the recordings of a specific team. *Data not included in the nuScenes format, as road cars following the autonomous race vehicle are visible in the sensor data.}
\label{tab:sensor_settings}
\begin{tabular}{lcccccccc}
\toprule
Team Name & GNSS & IMU & INS & Ground Speed & Wheel Speed & LiDAR & RADAR \\ 
\midrule
Code19 (C19) & \cmark & \cmark & \cmark & \cmark & \cmark & 20 Hz & 15 Hz \\ 
TU Munich (TUM) & \cmark & \cmark & \cmark & \cmark & \cmark & 10 Hz & 15 Hz \\ 
Polimove (POL) & \cmark & \cmark & \cmark & \cmark & \cmark & 20 Hz & 15 Hz \\ 
Unimore (UNI) & \cmark & \cmark & \cmark & & & 20 Hz & 15 Hz \\ 
Kinetiz (KIN*) & \cmark & \cmark & \cmark & & \cmark & 20 Hz & 15 Hz \\ 
Humda Labs (HUM*) & \cmark & \cmark & \cmark & \cmark & \cmark & 10 Hz & 15 Hz \\ 
Fly Eagle (FLY*) & & & & & & 10 Hz & \\ 
Constructur (CON) & \cmark & \cmark & \cmark & \cmark &  & 10 Hz &  15 Hz \\ 
\midrule
\end{tabular}
\end{table*}

\subsection{Racecar and Racetrack}

\begin{figure}[b]
    \centering
    \includegraphics[width=1\linewidth]{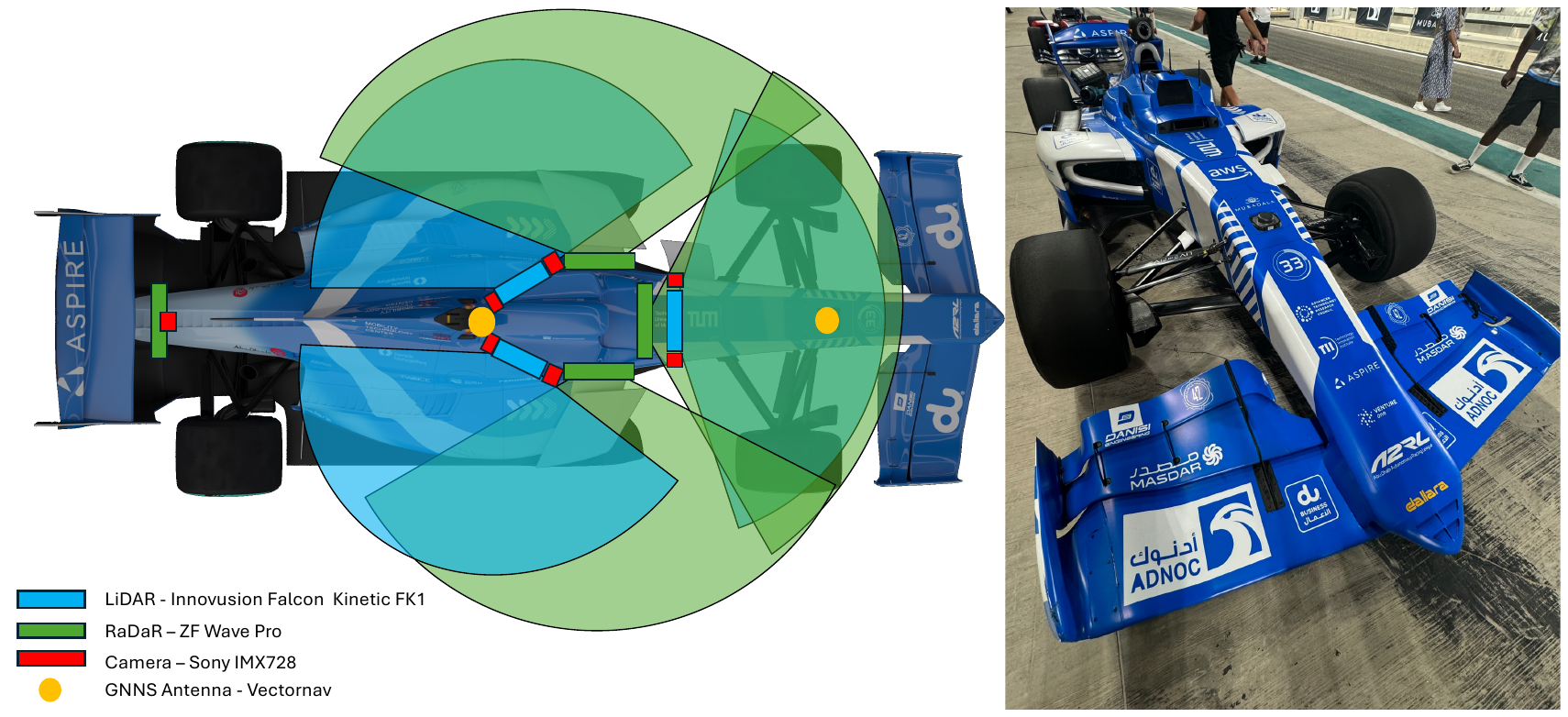}
    \caption{A2RL autonomous racing vehicle and sensor placement}
    \label{fig:eav24_sen}
\end{figure}

The A2RL racecar is built on a 2023 Dallara Super Formula chassis and is equipped with a turbocharged 2.0-liter four-cylinder engine generating approximately 550 horsepower.
The driver cockpit is replaced with the necessary hardware for autonomous driving, such as a steer-by-wire system and high-performance compute hardware based on an AMD EPYC 7313P CPU and an NVIDIA RTX 6000 ADA GPU. 
The car is equipped with the necessary sensors, and they are placed in the cockpit area where the driver usually sits (Figure \ref{fig:eav24_sen}). 

The vehicle is equipped with three LiDAR sensors (Seyond Falcon K), providing coverage of the front, rear-left, and rear-right regions. 
Although the LiDAR FoVs amount to a theoretical 360° surround view, the rear wing shades significant areas behind the car.
Furthermore, no LiDAR faces directly toward the rear, effectively limiting the view to roughly 60 meters behind the car, making rear detections especially difficult.
Four RADAR sensors (ZF ProWave 4D-RADAR) are mounted to offer full 360-degree coverage. Seven cameras (Leopard Imaging IMX728-9295-120H and -80H) are also installed to capture a complete 360-degree field of view around the vehicle. Furthermore, additional sensors are installed on the vehicle, including a dual-antenna INS (Vectornav VN-310) and a ground-speed sensor (Kistler SF Motion).
The GNSS integrated into the INS system uses NTRIP corrections from a base station near the racetrack, enabling RTK processing and yielding GNSS positional accuracy of approximately 3 to 5 cm.
All sessions were held at the Yas Marina Circuit Abu Dhabi, which features a total of 16 corners, including medium-speed turns, hard braking zones at turns 5 and 6, and high-speed corners such as turns 2 and 8.
In particular, the track provides significant localization challenges, as GNSS signals often fail in certain areas, such as the hotel section (turns 13 and 14) and under multiple bridges throughout the racetrack.
Figure \ref{fig:multifig} visualizes the sensor outputs from LiDARs and RADARs, along with the track layout and a trajectory around the track.

\subsection{Data Collection}

Teams are free to rely on only a subset of the sensors and can configure them according to their needs. 
As a result, teams use and record data from various sensors with different settings, leading to non-homogeneous data acquisition across teams. 
In addition to perception data, several teams provide INS and odometry.
Table \ref{tab:sensor_settings} provides a summary of the data contributed by each team.
The A2RL V\textsubscript{max} dataset consists of 297 scenes with 466 tracks of opponent vehicles. 
It features 305,531 LiDAR point clouds, of which 28,791 are annotated, totaling 38,545 vehicle bounding boxes.
386,006 RADAR point clouds were recorded alongside the LiDAR point clouds.




\subsection{Data Organization}

The data is provided in the nuScenes format. 
This format offers a significant advantage for perception evaluation, as the nuScenes devkit supports the execution and evaluation of tasks such as detection and tracking. 
Moreover, third-party libraries such as OpenPCDet \cite{OpenPCDet} natively support this format for training deep-learning models to detect 3D objects.
To facilitate seamless integration with existing tools, we aim for maximum consistency with the nuScenes dataset, ensuring that models trained on nuScenes can be used on our dataset with minimal modifications and evaluated using the nuScenes devkit.

In the nuScenes format, recordings are grouped into logs, with each log representing a recording of a specific location captured by a vehicle.
Every log consists of multiple scenes, each of which contains additional information, such as the ground truth of other race vehicles. 
A log is provided for each team and session.
Moreover, opponent perception tasks, such as detection, tracking, and prediction, only require frames in which opponents are visible. 
Hence, we only export scenes, where that is the case.
The scenes are shortened to less than 10 seconds when other vehicles are no longer in view.
The 3D data annotation is performed on LiDAR point clouds, using a single class, \textit{car}, that represents race vehicles.
In the nuScenes dataset, each bounding box is additionally annotated with a velocity, we do not provide this.
All frames consist of a single point cloud that is generated by combining measurements from all three LiDAR sensors. 
We chose this format because extrinsic calibration differs between team vehicles due to minor misalignment of the sensors and other factors such as installed rubber pads to decrease vibration.
Additionally, some teams recorded pointclouds only in an aggregated form, with no option to split them at a later point. 
All points are transformed into a right-hand coordinate system with its origin at the midpoint of the rear axle.

\section{Baseline Evaluation}
\label{sec:evaluation}

Our primary goal is to provide an easy-to-use, large-scale dataset for autonomous racing. 
However, to establish a reference point for assessing the dataset and methods used in high-speed autonomous racing perception, we provide a baseline evaluation for 3D object detection and tracking. The baseline evaluations further demonstrate the value of A2RL as a stress-testing environment for perception algorithms. The combination of extreme velocities, sparse long-range returns, and dynamic multi-vehicle interactions introduces an operational complexity that exceeds that of conventional autonomous driving datasets. As such, A2RL V\textsubscript{max} serves as a benchmark not only for algorithmic performance but also for system robustness under racing-grade constraints.

We do not reimplement well-known algorithms or conduct extensive parameter searches and tuning; instead, we apply them to our dataset to identify strengths and weaknesses in the available methods.
We want to emphasize future research and limit the evaluation in this work to off-the-shelf autonomous driving methods, with minor parameter adjustments to adapt to the provided dataset.

\subsection{Detection}
\label{sec:detection}

\begin{figure}[b]
    \centering
    \includegraphics[width=1\linewidth]{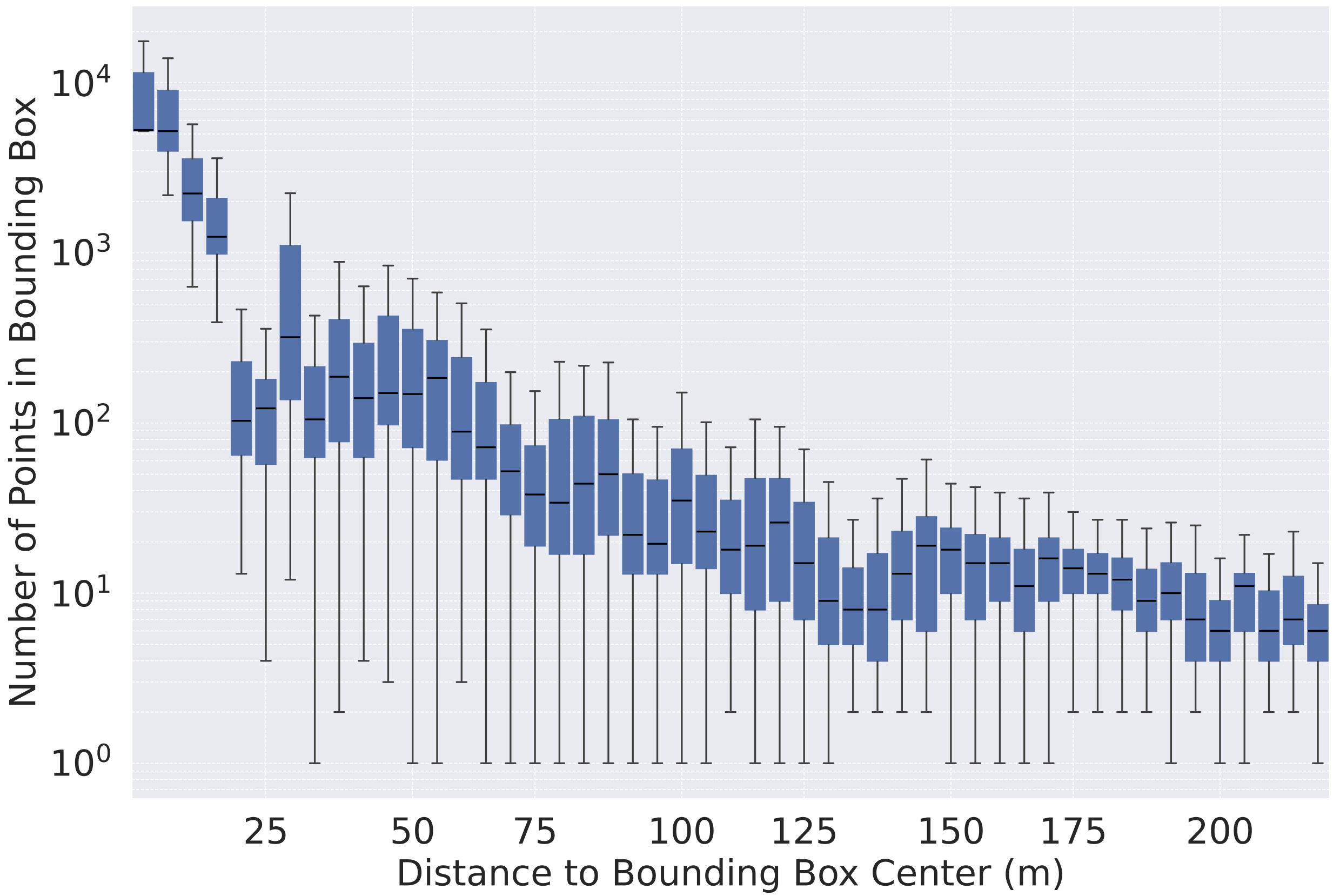}
    \caption{Relationship between the distance to an opponent vehicle and the number of LiDAR points within the annotated bounding box. The drop in point density between 15 and 25 meters is due to blind spots in close proximity.}
    \label{fig:points-distances}
\end{figure}

\begin{figure}[t]
    \centering
    \includegraphics[width=1\linewidth]{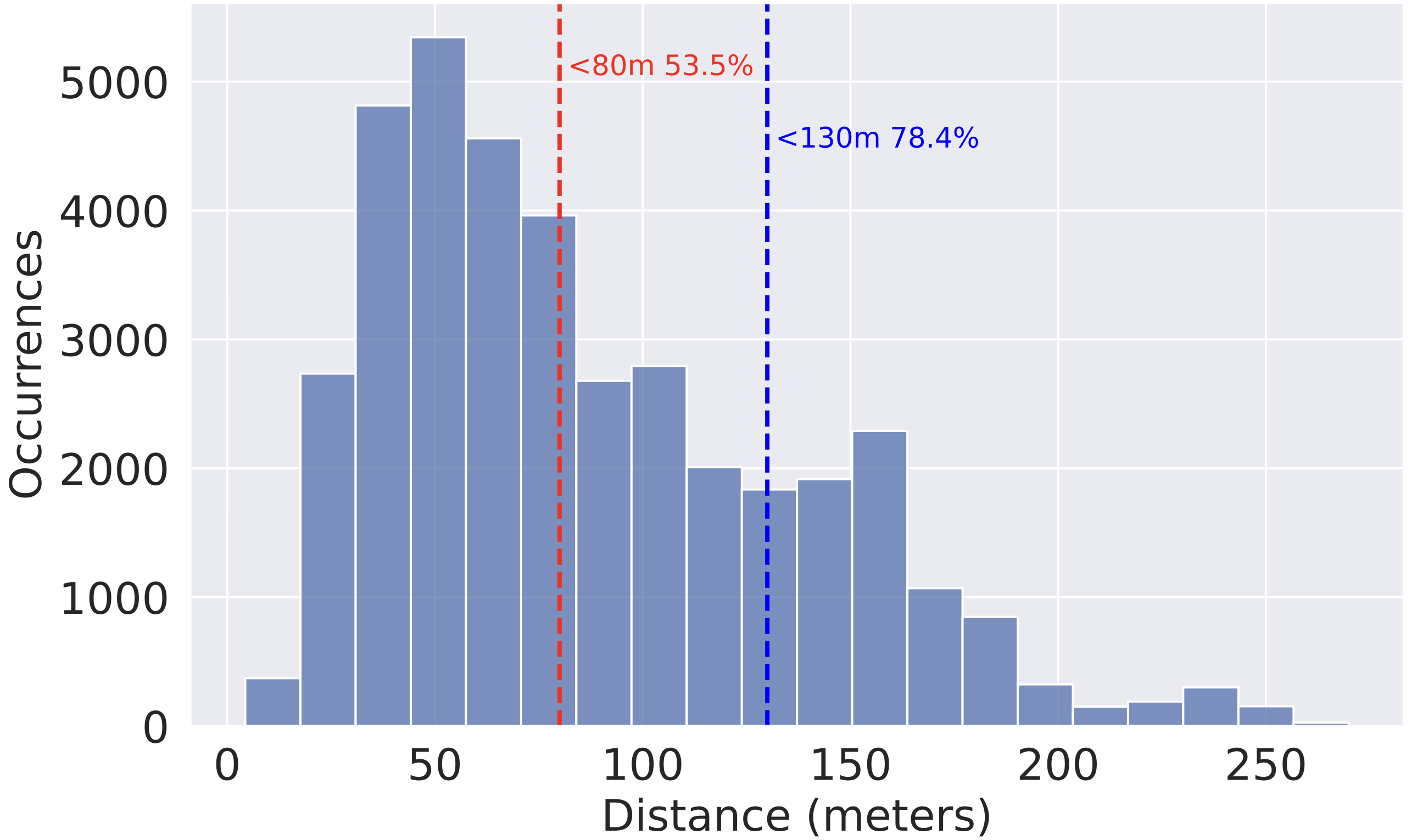}
    \caption{Distribution of Euclidean distances between the LIDAR sensor and annotated objects across all scenes in the dataset. Additionally, the red vertical line indicates the percentage of annotations within 80 meters, while the blue line marks those within 130 meters.}
    \label{fig:distance-to-opponents}
\end{figure}

There are two main challenges in this dataset that are not featured in other open datasets like nuScenes, Waymo or KITTI. 
First, the necessity to detect objects at ranges above 80 meters. This poses a challenge, as targets at long ranges return only a sparse set of LiDAR points. 
Figure \ref{fig:points-distances} quantifies this, showing the number of LiDAR points within annotated bounding boxes in relation to the distance to the bounding box.
The Figure reveals a steep decline in the number of LiDAR points per target up to approximately 80~m. 
Beyond this range, most targets contain fewer than 100 points, with those at 200~m receiving as few as 10 points per target.
Additionally, Figure \ref{fig:distance-to-opponents} highlights that 53.5$\%$ of the annotated data is within 80 meters, while 78.4$\%$ is within 130 meters. 
Hence, a substantial portion of annotations extends beyond 80 meters, emphasizing the need for detection methods capable of handling increasingly sparse reflections.

The second challenge is processing large point clouds (more than 150.000 points) in real time with low latency.
An efficient algorithm is necessary to process point clouds in less than 50 ms for the data captured at 20 Hz.
Detection latency is measured on the same GPU as used in the racecar, ensuring a realistic evaluation for real-world deployment.
We focus solely on the detection method's performance, noting that additional delays may arise from point cloud aggregation and data transfer.
Real-time evaluations often consider only a portion of the software stack; for example, if the detection method consumes 40 ms of the available 50 ms budget, subsequent processing stages may exceed the remaining 10 ms, potentially causing delays. 
Since there is no standardized approach to account for downstream latency, which depends on the methods used for prediction, planning, and controls, our discussion of execution time is qualitative.
In our evaluations, we define real-time capability as achieving a processing time of 30~ms or less.
Finally, we evaluate the 3D detection methods as originally published without further optimizations, such as those provided by TensorRT, which could drastically reduce the runtime of the methods.

Detections are evaluated using the detection task of the nuScenes evaluation toolkit. 
We focus on the Average Precision (AP) metric for the car class, as our dataset contains only a single class. Additionally, we evaluate using the Average Orientation Error (AOE) \cite{NEURIPS2024_947b6383} and the Average Translation Error (ATE) \cite{NEURIPS2024_947b6383}.
Since detection range is a key challenge, we modified the nuScenes evaluation toolkit to additionally report metrics based on the relative distance to the ego vehicle.
Hence, we evaluate detections across three distance ranges: 0~m to 80~m, 80~m to 130~m, and beyond 130~m.
To ensure robustness, each detection method is trained three times. We report only the mean of the metrics, as the standard deviation is negligible.

\begin{table*}
    \centering
    \caption{Evaluation of detection baseline performance. The best-performing detectors, as well as those that meet the latency requirement, are marked in bold.}
    \label{tab:detection_results}
    \resizebox{\textwidth}{!}{ 
    \begin{tabular}{l c | ccc | ccc | ccc | c}
        \toprule
        \multirow{2}{*}{Method} & \multirow{2}{*}{Publication} & \multicolumn{3}{c|}{AP (\%) \small ↑} & \multicolumn{3}{c|}{AOE (rad) \small ↓} & \multicolumn{3}{c|}{ATE (m) \small ↓} & \multirow{2}{*}{Latency (ms) \small ↓} \\
        & & 0-80~m & 80-130~m & 130+~m & 0-80~m & 80-130~m & 130+~m & 0-80~m & 80-130~m & 130+~m & \\
        \midrule
        SECOND \cite{second} & Sensors 2018 & 0.716 & 0.583 & 0.257 & 0.179 & 0.193 & 0.272 & 0.094 & 0.102 & 0.091 & \textbf{27.7 $\pm$ 0.3} \\
        PointPillars \cite{8954311} & CVPR 19 & 0.755 & 0.594 & 0.281 &  0.165 & 0.178 & 0.257 & 0.104 & 0.090 & 0.099 & 52.2 $\pm$ 0.5 \\
        PV-RCNN \cite{shi2020pv} & CVPR 20 & 0.809 & 0.562 & 0.201 & \textbf{0.158} & 0.175 & 0.253 & 0.086 & 0.093 & 0.090 & 185.8 $\pm$ 0.3 \\
        Centerpoint (Pillar) \cite{Yin_2021_CVPR} & CVPR 21 & 0.779 & 0.611 & 0.324 & 0.172 & 0.190 & 0.288 & 0.116 & 0.161 & 0.149 & 34.2 $\pm$ 0.5 \\
        Centerpoint (Voxel) \cite{Yin_2021_CVPR} & CVPR 21 & 0.769 & \textbf{0.620} & \textbf{0.329} & 0.180 & 0.191 & 0.278 & 0.124 & 0.105 & 0.125 & 33.6 $\pm$ 0.2 \\
        PV-RCNN++ \cite{Shi2023} & IJCV 23 & \textbf{0.812} & 0.519 & 0.135 & 0.157 & \textbf{0.171} & \textbf{0.238} & \textbf{0.079} & \textbf{0.087} & \textbf{0.084} & 69.9 $\pm$ 0.3 \\
        DSVT (Pillar) \cite{wang2023dsvt} & CVPR 23 & 0.753 & 0.581 & - & 0.172& 0.192 & - & 0.097 & 0.095 & - & 73.6 $\pm$ 0.1 \\
        DSVT (Voxel) \cite{wang2023dsvt} & CVPR 23 & 0.752 & 0.470 & - & 0.168 & 0.190 & - & 0.098 & 0.090 & - & 75.6 $\pm$ 0.2 \\
        Voxel-Mamba \cite{NEURIPS2024_947b6383} & NeurIPS 24 & 0.765 & 0.582 & - & 0.182 & 0.207 & - & 0.122 & 0.089 & & 86.2 $\pm$ 0.2 \\
        VoxelNeXt \cite{chen2023voxenext} & CVPR 24 & 0.805 & 0.609 & 0.301 & 0.180 & 0.214 & 0.294 & 0.117 & 0.136 & 0.148 & 36.3 $\pm$ 0.4 \\
        LION \cite{liu2024lion}  & NeurIPS 24 & 0.780 & 0.593 & 0.226 & 0.184 & 0.218 & 0.295 & 0.176 & 0.214 & 0.129 & 116.2 $\pm$ 0.2 \\
        \bottomrule
    \end{tabular}
    }
\end{table*}

The detection results are summarized in Table \ref{tab:detection_results}.
In terms of Average Precision (AP), all detection methods achieve a similar performance within the 0–80~m range, to the performance of the car class in their original publications. 
In particular, more recent methods outperform early 3D detection approaches, with SECOND achieving an AP of approximately 0.7, while state-of-the-art methods reach around 0.8.
However, in higher detection ranges 80–130~m and beyond 130~m, these improvements diminish, with little to no advancement over past methods. 
Moreover, detection performance drops significantly at higher distances.
This highlights the limitations of current methods in long-range perception and reinforces the need for specialized approaches to improve detection beyond 80 meters.
Notably, DSVT and Voxel-Mamba could only be trained successfully with a square field of view, which required extensive GPU memory. 
Consequently, we restricted their field of view to -130~m to 130~m to ensure feasible training.
All methods achieve a similar AOE of approximately 0.18 radians, corresponding to a mean orientation error of around 10 degrees. 
At greater distances, the AOE remains consistent across methods, increasing to approximately 0.27 radians (16 degrees).
Similar to Average Precision, this error increases with distance to the target, reaching values up to 0.28 radians (approximately 15 degrees) at longer ranges. 
However, we do not consider this a critical issue, as AOE is most crucial in close proximity, such as during head-to-head interactions.
The ATE is  $\sim$0.1m  in all methods.
In contrast, the ATE remains consistent across all distance ranges and is relatively accurate, indicating that objects are correctly localized when detected.
We attribute this to the fact that all ground-truth objects have identical dimensions, which reduces variability in localization errors and ensures stable performance across different detection ranges.

During inference evaluation, we set the batch size to 1. 
To ensure consistent timing measurements, we first perform a single inference pass over the validation split to warm up the GPU. 
Subsequently, we iterate over the validation split five times, reporting the mean execution time per sample along with the standard deviation.
Only the SECOND method is fast enough based on our conservatively set runtime threshold.
Methods that mainly rely on sparse convolution in their backbone, such as PointPillars, CenterPoint, and VoxelNeXt, achieve runtimes close to the conservatively set threshold.
Unfortunately, DSVT does not meet the runtime requirements, despite being designed with efficiency in mind. 
The primary reason for this is that its inference time scales with the field of view, making it computationally demanding for larger scenes.
In contrast, linear RNN-based methods, including Voxel-Mamba and LION, are the slowest among the evaluated approaches and would require extensive optimization to meet real-time constraints.

We further evaluated whether VoxelNext detectors trained on standard autonomous-driving datasets generalize to racing. Detectors trained on nuScenes and Argoverse achieved near-zero AP on the A2RL V\textsubscript{max}, revealing a substantial domain gap between urban driving and autonomous racing.


\subsection{Tracking}


We evaluate object tracking using the nuScenes devkit. 
The evaluation follows a tracking-by-detection paradigm, where objects are first detected using an object detection method and then tracked by a tracking algorithm. 
As a baseline, the nuScenes benchmark employs the AB3DMOT \cite{Weng2020_AB3DMOT} tracking algorithm. 
We follow this approach and provide a tracking baseline using the same algorithm.
In our evaluation, we solely focus on the evaluation of the tracking algorithm, this is achieved by using perfect detections.
This serves as an upper bound, isolating the tracking algorithm's performance from any detection errors.


\begin{table*}
\centering
\caption{Tracking evaluation results. In the default configuration, tracks are only matched to detections if they are within 2 meters distance.}
\label{tab:tracking_results}
\begin{tabular}{lcccccccc}
\toprule
Detection Method & Max Range & AMOTA $\uparrow$ & AMOTP $\downarrow$ & IDS $\downarrow$ & FRAG $\downarrow$ & FP $\downarrow$ & FN $\downarrow$ & Latency $\downarrow$ \\
\midrule
Perfect detections (default) & 300~m & 0.814 & 0.357 & 260 & 336 & 2178 & 3732 & 0.3ms $\pm$ 0.4 \\
\midrule
Perfect detections & 300~m & 0.891 & 0.303 & 83 & 291 & 1432 & 2559 & " \\
Perfect detections & 130~m & 0.897 & 0.292 & 61 & 177 & 910 & 1749 & " \\
Perfect detections & 80~m & 0.929 & 0.223 & 32 & 81 & 455 & 952 & " \\
\bottomrule
\end{tabular}
\end{table*}

The official implementation of AB3DMOT\footnote{\url{https://github.com/xinshuoweng/AB3DMOT}} provides multiple configurations depending on the tracking target and the dataset. 
We aim to keep the evaluation as close as possible to the original nuScenes dataset. 
Hence, we initialize the tracker using the same configuration as the car class in the nuScenes dataset.
However, minor adjustments are made to improve overall tracking performance.
Since not all teams used cameras and thus did not provide camera calibration, we use the Euclidean distance between 3D boxes as the distance metric, rather than relying on projected bounding boxes in camera space.
To reduce identity switches, we change the maximum matching distance from 2 meters to 8 meters between tracks and detections, as racecars travel further between frames.
Note that the Kalman filter within AB3DMOT estimates the velocity for each track; however, as our ground truth annotations do not include velocity information, we set the predicted velocity to 0 during evaluation.
An interesting observation for tracking is that relatively few racecars need to be tracked per frame.
Specifically, a single racecar is present in 18,137 frames, two racecars in 8,086 frames, and three opponent racecars in 1,412 frames.
This is primarily due to the significant performance gaps between teams during the race, which lead to large separations between vehicles.
Consequently, the data-association aspect of multi-object tracking is simplified, as only a small number of cars need to be tracked at any given time.
However, state estimation of opponent tracks is significantly harder as differential speeds are far greater than in regular vehicle tracking.
Our evaluation is primarily based on the AMOTA, AMOTP, IDS, and FRAG metrics. \newline
Table \ref{tab:tracking_results} presents the evaluation results. 
Surprisingly, the baseline method demonstrates strong tracking performance across all metrics. 
Based on the AMOTA metric, the tracker successfully detects and tracks objects most of the time, with errors occurring only in specific scenarios. 
Moreover, the low AMOTP indicates that the tracker's bounding boxes are well aligned with the ground truth.
However, the baseline leaves room for improvement, particularly regarding identity switches and fragmentation.
Tracking is relatively simple on straight road segments, as the longitudinal velocity between the ego vehicle and the tracked object often varies alone.
However, in high-speed corners, tracking becomes considerably more challenging, introducing a high number of fragmentations and identity switches.
Rapid changes in the ego vehicle's yaw induce significant movement in tracked objects, leading to track fragmentation.
This is especially true for tracked objects far from the ego-vehicle.
We reason that the Kalman filter used by the AB3DMOT algorithm is unable to track these quick changes, leading to identity switches and fragmentation.
To gain deeper insights into how distance affects the tracker's performance, we evaluate tracking based on the maximum range to a track.
However, in contrast to detection, we include all tracks up to the specified range.
This evaluation clearly shows that the range plays a key role in both fragmentation and identity switches.
Although tracks within 80 meters account for 53.5$\%$ of the data, both the IDS and FRAG metrics remain relatively low.
When the range is extended to 130 meters, IDS and FRAG almost double, despite the fact that the volume of data increases only by 20$\%$.
A similar trend is observed when including tracks up to the maximum range, further emphasizing the
need for more robust tracking methods that can effectively handle long-range scenarios and mitigate fragmentation and identity switches.
The tracker's runtime is negligible, taking less than one millisecond per frame.

\section{Discussion \& Conclusion}
\label{sec:conclusion}

We present an extensive dataset that contains professionally annotated LiDAR data and supplementary sensor data such as RADAR, GNSS, and additional vehicle sensor data.
To support future research on high-speed perception, the dataset is structured in an easily accessible format using the nuScenes development kit, enabling standardized benchmarking for perception and tracking.

Initial experiments with off-the-shelf baseline methods for 3D LiDAR detection yield promising results.  To evaluate their performance, we trained and tested well-established 3D detection models on the proposed A2RL V\textsubscript{max} dataset.
Although these methods accurately detect vehicles at close ranges, their performance drops significantly at long distances. 
Notably, existing 3D detection approaches have not been extensively studied for high detection ranges, and our results confirm that they fail to generalize in this domain.
This highlights a critical research gap in long-range 3D detection, which is required particularly in high-speed autonomous driving scenarios.
Following the approach of the nuScenes dataset, we evaluated AB3DMOT on our proposed dataset for tracking scenarios.
While the baseline yields promising results, it fails frequently in sharp movements, which are common in racing, leading to high track fragmentation

This dataset provides comprehensive evaluation tools for both detection and tracking.
Since the dataset provides annotations for the entire race, we plan to integrate motion forecasting in future iterations of the dataset. This is particularly crucial for head-to-head racing, where accurately predicting an opponent’s maneuvers enables more responsive and strategic decision-making. Moreover, the track poses significant localization challenges, as GNSS signals are often disrupted by bridges and tunnels. To address this, we plan to enhance the dataset with high-precision ground truth positioning, enabling more robust localization research for high-speed autonomous driving.

Beyond introducing a dataset, this work demonstrates how competition-driven research ecosystems can generate high-value scientific artifacts. Through ASPIRE’s strategic orchestration of A2RL, perception data was captured under controlled yet extreme conditions that are unattainable in conventional road-driving datasets. The resulting dataset embodies a new paradigm in autonomous systems research, where competition, benchmarking, and open scientific dissemination are intentionally integrate.

\bibliographystyle{IEEEtran}
\bibliography{main}

\end{document}